# A CONSTRAINT - PROPAGATION APPROACH
# TO PROBABILISTIC REASONING *

## Judea Pearl
## Computer Science Department, University of California, Los Angeles

## ABSTRACT

The paper demonstrates that strict adherence to probability theory does not preclude the use of concurrent, self-activated constraint-propagation mechanisms for managing uncertainty. Maintaining local records of sources-of-belief allows both predictive and diagnostic inferences to be activated simultanously and propagate harmoniously towards a stable equillibrium.

## 1. INTRODUCTION: BAYES NETWORKS AND CONSTRAINTS PROPAGATION

Scholarly textbooks on probability theory often create the impression that to construct an adequate representation of probabilistic knowledge we must first define a joint distribution function on all propositions and their combinations, i.e., on the so-called universe of discernment. The computational difficulties involved in articulating, validating, storing and manipulating such distributions seem insurmountable and have discouraged many AI researchers from openly using probabilistic formalisms in expert systems. In truth, however, these difficulties are merely mathematical fiction, and do not plague common-sensical approaches to probabilistic reasoning. In a sparsely connected world like ours, it is fairly clear that probabilistic knowledge, in both man and machine, should not be represented by entries of a giant joint-distribution table, but rather by a network of low-order probabilistic relationships between small clusters of semantically related propositions. One effective representation of such relationships is provided by *Bayesian Networks*: a class of networks typified by the use of "influence diagrams" in decision analysis [Howard and Matheson, 1984] and "inference networks" in expert systems [Duda, Hart, and Nilsson, 1976]. (The alternative network representation using Markov fields [Pearl, 1985] will not be discussed here.)

Bayes Networks are directed acyclic graphs in which the nodes represent propositions (or variables), the arcs signify the existence of direct causal influences between the linked propositions, and the strengths of these influences are quantified by conditional probabilities (Figure 1). Thus, if the graph contains the variables $x_1, \ldots, x_n$, and $S_i$ is the set of parents for variable $x_i$, then a complete and consistent quantification can be attained by specifying, for each node $x_i$, an assessment of $P(x_i \mid S_i)$. The product of all these assessments,

$$P(x_1, \ldots, x_n) = \prod_i P(x_i|S_i) \tag{1}$$

constitutes a joint-probability model which supports the assessed quantities. That is, if we compute the conditional probabilities $P(x_i \mid S_i)$ dictated by $P(x_1, \ldots, x_n)$, the original assessments are recovered. For example, the distribution corresponding to the graph of Figure 1 can be written by inspection:

---

* This work was supported in part by NSF Grant DSR 83-13875

$$P(x_1, x_2, x_3, x_4, x_5, x_6) = P(x_6|x_5) \, P(x_5|x_2, x_3) \, P(x_4|x_1, x_2) \, P(x_3|x_1) \, P(x_2|x_1) \, P(x_1).$$

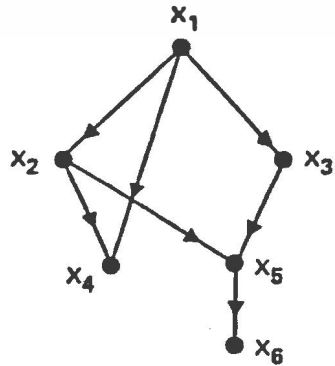

*Figure 1*

An important feature of Bayes network is that it provides a clear visual representation for many independence relationships embedded in the underlying probabilistic model. The criterion for detecting these independencies is based on graph separation: namely, if all paths between $x_i$ and $x_j$ are "blocked" by a a subset $S$ of variables, then $x_i$ is independent of $x_j$ given the values of the variables in $S$. Thus, each variable $x_i$ is independent of both its siblings and its grandparents, given the values of the variables in its parent set $S_i$. For this "blocking" criterion to hold in general, we must provide a special interpretation of separation for nodes that share common children. We say that the pathway along arrows meeting head-to-head at node $x_k$ is normally "blocked", unless $x_k$ or any of its descendants is in $S$. In Figure 1, for example, $x_2$ and $x_3$ are independent given $S_1 = \{x_1\}$ or $S_2 = \{x_1, x_4\}$, because the two paths between $x_2$ and $x_3$ are blocked by either one of these sets. However, $x_2$ and $x_3$ may not be independent given $S_3 = \{x_1, x_6\}$, because $x_6$, as a descendant of $x_5$, "unblocks" the head-to-head connection at $x_5$, thus opening a pathway between $x_2$ and $x_3$.

Once a Bayes network is established, it can be used to represent the deep causal knowledge of a domain expert and can provide probabilistic answers to all queries regarding the interpretation of evidential information in that domain. Ideally, however, we would also like to treat such a network as a computational architecture that facilitates the interpretation of data *at the knowledge level itself*; we want to view the links not merely as codes for storing factual knowledge but also as pathways and activation centers which both direct and propel the flow of data in the process of querying and updating that knowledge.

The process of self-activated interpretation is conveniently described in terms of constraint-propagation or relaxation paradigm. Each link in the network represents constraints on the possible values that the belief parameters can take at the two nodes connected by the link. Updating is accomplished by successively finding unsatisfied constraints and satisfying them by modifying the belief parameters, thus bringing "out of kilter" constraints back to relaxed status. Relaxing one constraint usually perturbs its neighbors, so relaxation results in a multi-directional propagation process which reaches a static equilibrium when all constraints are satisfied.

The relaxation paradigm has several advantages over other mechanisms of uncertainty management. It permits knowledge to be specified declaratively without regard for the specific control method used. It is readily implementable by pattern-oriented rule-based languages as well as by object-oriented languages. In the former, the antecedents of the rules alert to violations of constraints and their consequent parts specify corrective actions. In the latter, each node is an object of the same generic type and the constraints are the messages by which neighboring objects communicate. Moreover, relaxation can be executed in parallel by a large array of simple autonomous processes, thus providing a reasonable model of human cognitive behavior.

While constraint-propagation mechanisms have found several applications in AI, such as vision [Rosenfeld, Hummel and Zucker, 1976; Waltz, 1972] and truth maintenance [McAllester, 1980], their use in evidential reasoning has been limited to non-Bayesian formalisms [e.g. Lowrance, 1982]. The reason has been several-fold.

First, the conditional probabilities characterizing the links in the network do not seem to impose definitive constraints on the probabilities that can be assigned to the nodes. The quantifier $P(A|B)$ only restricts the belief accorded to $A$ in a very special set of circumstances: namely, when $B$ is known to be true with absolute certainty, and when no other evidential data is available. Under normal circumstances, all internal nodes in the network will be subject to some uncertainty and, more seriously, after observing evidence $e$ the conditional belief in $A$ is no longer governed by $P(A|B)$ but by $P(A|B, e)$, which may be totally different. The result is that any assignment of beliefs, $P(A)$ and $P(B)$, to propositions $A$ and $B$ can be consistent with the value of $P(A|B)$ initially assigned to the link connecting them; therefore, no violation of constraint can be detected locally.

Next, the difference between $P(A|B, e)$ and $P(A|B)$ seems to suggest that the weights on the links should not remain fixed but should undergo constant adjustment as new evidence arrives. This, in turn, would require an enormous computational work and would wipe out the advantages normally associated with propagation through fixed constraints.

Finally, the fact that evidential reasoning involves both top-down (predictive) and bottom-up (diagnostic) inferences has caused apprehensions that, once we allow the propagation process to run its course unsupervised, pathological cases of instability, deadlock, and circular reasoning will develop. Indeed, if a stronger belief in a given hypothesis means a greater expectation for the occurrence of its various manifestations and if, in turn, a greater certainty in the occurrence of these manifestations adds further credence to the hypothesis, how can one avoid infinite updating loops when the processors responsible for these propositions begin to communicate with one another? Likewise, Lowrance [1982] expresses concern that if proposition $B$ influences the belief in $A$ via $P(A|B)$ and proposition $A$ influences the belief in $B$ via $P(B|A)$, then the "feedback between $A$ and $B$ would eventually drive the two beliefs to the marginals," thus preventing any further updating from occurring.

This paper shows that coherent and stable probabilistic reasoning can be accomplished by local propagation mechanisms while keeping the weights on the links constant throughout the process. This is made possible by characterizing the belief in each proposition by a vector of parameters, one for each port. Each component in the vector stands for the degree of support that the host proposition obtains from one of its neighbors. We show that, in certain networks, maintaining such a breakdown record of the sources of belief facilitates efficient updat-The relaxation paradigm has several advantages over other mechanisms of uncertainty management. It permits knowledge to be specified declaratively without regard for the specific control method used. It is readily implementable by pattern-oriented rule-based languages as well as by object-oriented languages. In the former, the antecedents of the rules alert to violations of constraints and their consequent parts specify corrective actions. In the latter, each node is an object of the same generic type and the constraints are the messages by which neighboring objects communicate. Moreover, relaxation can be executed in parallel by a large array of simple autonomous processes, thus providing a reasonable model of human cognitive behavior.

While constraint-propagation mechanisms have found several applications in AI, such as vision [Rosenfeld, Hummel and Zucker, 1976; Waltz, 1972] and truth maintenance [McAllester, 1980], their use in evidential reasoning has been limited to non-Bayesian formalisms [e.g. Lowrance, 1982]. The reason has been several-fold.

First, the conditional probabilities characterizing the links in the network do not seem to impose definitive constraints on the probabilities that can be assigned to the nodes. The quantifier $P(A|B)$ only restricts the belief accorded to $A$ in a very special set of circumstances: namely, when $B$ is known to be true with absolute certainty, and when no other evidential data is available. Under normal circumstances, all internal nodes in the network will be subject to some uncertainty and, more seriously, after observing evidence $e$ the conditional belief in $A$ is no longer governed by $P(A|B)$ but by $P(A|B, e)$, which may be totally different. The result is that any assignment of beliefs, $P(A)$ and $P(B)$, to propositions $A$ and $B$ can be consistent with the value of $P(A|B)$ initially assigned to the link connecting them; therefore, no violation of constraint can be detected locally.

Next, the difference between $P(A|B, e)$ and $P(A|B)$ seems to suggest that the weights on the links should not remain fixed but should undergo constant adjustment as new evidence arrives. This, in turn, would require an enormous computational work and would wipe out the advantages normally associated with propagation through fixed constraints.

Finally, the fact that evidential reasoning involves both top-down (predictive) and bottom-up (diagnostic) inferences has caused apprehensions that, once we allow the propagation process to run its course unsupervised, pathological cases of instability, deadlock, and circular reasoning will develop. Indeed, if a stronger belief in a given hypothesis means a greater expectation for the occurrence of its various manifestations and if, in turn, a greater certainty in the occurrence of these manifestations adds further credence to the hypothesis, how can one avoid infinite updating loops when the processors responsible for these propositions begin to communicate with one another? Likewise, Lowrance [1982] expresses concern that if proposition $B$ influences the belief in $A$ via $P(A|B)$ and proposition $A$ influences the belief in $B$ via $P(B|A)$, then the "feedback between $A$ and $B$ would eventually drive the two beliefs to the marginals," thus preventing any further updating from occurring.

This paper shows that coherent and stable probabilistic reasoning can be accomplished by local propagation mechanisms while keeping the weights on the links constant throughout the process. This is made possible by characterizing the belief in each proposition by a vector of parameters, one for each port. Each component in the vector stands for the degree of support that the host proposition obtains from one of its neighbors. We show that, in certain networks, maintaining such a breakdown record of the sources of belief facilitates efficient updat-ing of parameters by constraint-propagation, and that the network relaxes to a stable equilibrium consistent with the axioms of probability theory, in time proportional to the network diameter. This record of parameters is also postulated as the mechanism which permits people to trace back the sources of beliefs for the purpose of constructing explanatory arguments.

## 2. *PROPAGATION IN SINGLY-CONNECTED NETWORKS*

We shall first consider Bayes networks which are singly connected, that is, there is at most one underlying path between any pair of nodes. Propagation algorithms for such networks were developed by Pearl [1982], for the special case of trees, and were later generalized by Kim and Pearl [1983] to admit nodes with multiple parents. To establish the notation necessary for treating more general networks, we shall reiterate here the results of Kim and Pearl and cast them in the context of constraint propagation.

Let each node in the network represent a multivalued variable which might stand for a collection of mutually exclusive hypotheses (e.g., identity of organism: $ORG_1$, $ORG_2$,...) or a collection of possible observations (e.g. patient's temperature: high, medium, low). Let a variable be labeled by a capital letter, e.g., $A, B, C, ...$, and its possible values subscripted, e.g., $A_1, A_2, ..., A_n$. Each group of arrows pointing at a given node is quantified by a fixed conditional probability matrix. For example, the arrows $B \rightarrow A$ and $C \rightarrow A$ in Figure 2, will be quantified by a matrix $M$, with entries: $M_{ijk} = P(A_i|B_j, C_k)$.

These matrices quantify the strength of influence between causes and their consequences. Additionally, they contain the information for deciding how the belief in one cause is affected by evidence bearing on another, once their common manifestation is observed. This interaction, colloquially termed "explaining away," is a prevailing pattern of human reasoning, and occurs even when the causal variables are marginally independent. For example, when a physician discovers evidence in favor of one disease, it reduces the credibility of other diseases, although the patient could be suffering from two or more disorders simultaneously.

Instantiated variables, constituting the incoming evidence or **data** will be denoted by $D$. For the sake of clarity we will distinguish between the fixed conditional probabilities that label the links, e.g. $P(A|B)$, and the dynamic values of the updated node probabilities. The latter will be denoted by $BEL(A_i)$, which reflects the overall belief accorded to proposition $A_i$ by all data so far received. Thus,

$$BEL(A_i) \overset{\Delta}{=} P(A_i|D) \qquad (2)$$

where $D$ is the value combination of all instantiated variables.

### Fusion Equations

Consider a fragment of a singly connected network, as depicted in Figure 2. The link $B \rightarrow A$ partitions the graph into two parts: an upper subgraph $G_{BA}^+$, and a lower subgraph $G_{BA}^-$, the complement of $G_{BA}^+$. These two graphs contain two sets of **data** which we shall call $D_{BA}^+$ and $D_{BA}^-$, respectively. Likewise, the links $C \rightarrow A$, $A \rightarrow X$, and $A \rightarrow Y$ define the subgraphs $G_{CA}^+$, $G_{AX}^-$, and $G_{AY}^-$ which contain the data sets $D_{CA}^+$, $D_{AX}^-$, and $D_{AY}^-$, respectively. Since $A$ is a common child of $B$ and $C$, it does not separate $G_{BA}^+$ and $G_{CA}^+$ apart. However, it does separate the following three subgraphs: $G_{BA}^+ \bigcup G_{CA}^+$, $G_{AX}^-$, and $G_{AY}^-$, and we can write

$$P(D_{BA}^+, D_{CA}^+, D_{AX}^-, D_{AY}^-|A_i) = P(D_{BA}^+, D_{CA}^+|A_i) \, P(D_{AX}^-|A_i) \, P(D_{AY}^-|A_i) \tag{3}$$

Thus, using Bayes rule, the overall strength of belief in $A_i$ can be written:

$$\text{BEL}(A_i) = P(A_i|D_{BA}^+, D_{CA}^+, D_{AX}^-, D_{AY}^-) = \alpha \, P(A_i|D_{BA}^+, D_{CA}^+) \, P(D_{AX}^-|A_i) \, P(D_{AY}^-|A_i) \tag{4}$$

where $\alpha$ is a normalizing constant. By further conditioning over the values of $B$ and $C$, we get

$$\text{BEL}(A_i) = \alpha \, P(D_{AX}^-|A_i) \, P(D_{AY}^-|A_i) [\sum_{jk} P(A_i|B_j C_k) \, P(B_j|D_{BA}^+) \, P(C_k|D_{CA}^+)]..br \tag{5}$$

Eq.(5) shows that the probability distribution of each variable $A$ in the network can be comput-
ed if three types of parameters are made available: (1) the current strength of the causal sup-
port, $\pi$, contributed by each incoming link to $A$;

$$\pi_A(B_j) = P(B_j|D_{BA}^+) \tag{6}$$

(2) the current strength of the diagnostic support, $\lambda$, contributed by each outgoing link from $A$;

$$\lambda_X(A_i) = P(D_{AX}^-)|A_i) \tag{7}$$

and (3) the fixed conditional probability matrix, $P(A|B, C)$, which relates the variable $A$ to its
immediate causes. Accordingly, we let each link carry two dynamic parameters, $\pi$ and $\lambda$, and
let each node store the information contained in $P(A|B,C)$.

With these parameters at hand, the fusion equation (5) becomes

$$\text{BEL}(A_i) = \alpha \, \lambda_X(A_i) \, \lambda_Y(A_i) \sum_{jk} P(A_i|B_j C_k) \, \pi_A(B_j) \, \pi_A(C_k) \tag{8}$$

Alternatively, from two parameters, $\pi$ and $\lambda$, residing on the same link we can compute the be-
lief distribution of the parent node by the product

$$\text{BEL}(B_j) = \alpha \, \pi_A(B_j) \, \lambda_A(B_j) \tag{9}$$

### Updating Equations

Assume that the vectors $\pi$ and $\lambda$ are stored with each link, $\pi$ at the tail of the arrow
and $\lambda$ at its head. Our task is now to prescribe how the values of $\pi$ and $\lambda$ at a given link are
constrained by the corresponding parameters at neighboring links.

*Updating* $\lambda$: Starting from the definition of $\lambda_A(B_i) = P(D_{BA}^-|B_i)$, we partition the data $D_{BA}^-$ into its
components: $A$, $D_{AX}^-$, $D_{AY}^-$, $D_{CA}^+$, and summing over all values of $A$ and $C$ we get

$$\lambda_A(B_i) = \alpha \sum_j [\pi_A(C_j) \sum_k \lambda_X(A_k) \, \lambda_Y(A_k) \, P(A_k|B_i C_j)]. \tag{10}$$

Eq.(10) shows that only three parameters (in addition to the conditional probabilities $P(A|B, C)$)
need to be involved in updating the diagnostic parameter vector $\lambda_A(B)$: $\pi_A(C)$, $\lambda_X(A)$, and $\lambda_Y(A)$.
This is expected since $D_{BA}^-$ is completely summarized by $X$, $Y$, and $C$.

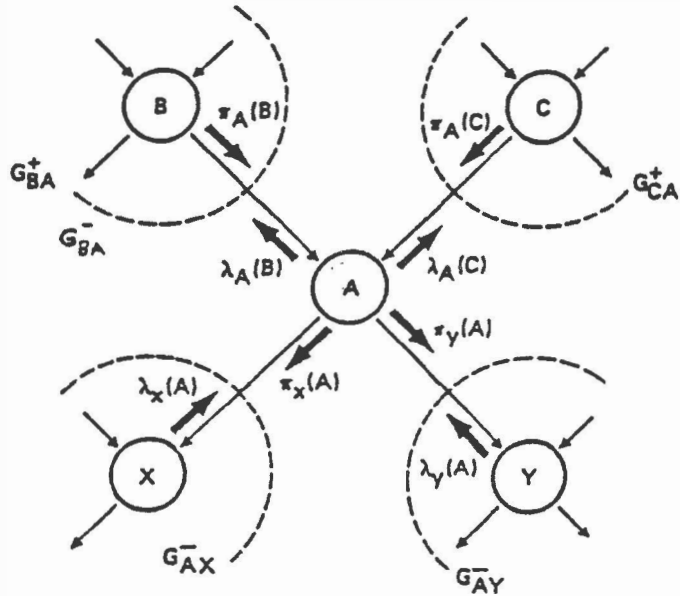

*Figure 2*

*Updating* $\pi$: The rule for updating the causal parameter $\pi_X(A)$ is governed the formula:

$$\pi_X(A_i) = \alpha\lambda_Y(A_i)[\sum_{jk}P(A_i|B_jC_k)\ \pi_A(B_j)\pi_A(C_k)] \tag{11}$$

Thus, $\pi_X(A)$, like $\lambda_A(B)$, is also determined by three neighboring parameters: $\lambda_Y(A)$, $\pi_A(B)$, and $\pi_A(C)$.

The boundary conditions are established as follows:

1. *Data-nodes*: If the $j^{th}$ state of $B$ is known to be true, we add to $B$ a dummy son $Z$, and set $\lambda_Z(B) = (0,...,0,1,0,...,0)$ with 1 at the $j^{th}$ position.

2. *Anticipatory nodes*: If $B$ is a childless node that has not been instantiated, we set $\lambda(B) = (1,1,...,1)$.

3. *Root-nodes*: If $B$ is a node with no parents, we add to $B$ a dummy father $Z$ instantiated to $Z=TRUE$, and set the link matrix $P(B|Z=TRUE)$ equal to the prior $P(B)$.

## Constraints Propagation

So far we have viewed the links of the network as message-carrying devices through which the node processors communicate. They can also be viewed as constraint-maintaining agents. Imagine that each node is characterized by several parameter vectors, one for each of its ports. The $\pi$'s are placed on the outgoing ports and the $\lambda$'s on the incoming ports. In node $A$ of Figure 2, for example, the parameters $\pi_X(A)$ and $\pi_Y(A)$ will be placed on the lower two ports (facing the children $X$ and $Y$) while $\lambda_A(B)$ and $\lambda_A(C)$ will be placed on the upper ports (facing the parents). Imagine also that a node is permitted to compare its own set of parameters with those of its neighbors. Equations (10) and (11) then dictate how the link matrices $P(A|B,C)$ impose equality constraints between the parameters of one node and those of its neighbors. If these equalities are satisfied (within some reasonable tolerance), no activity takes place. However, if any of these equalities is violated, the responsible node is activated to revise its violating parameter and set it straight. This, of course, will activate similar revisions at neighboring nodes and will set up a multidirectional propagation process.

Eqs.(10) and (11) demonstrate that a perturbation of the causal parameter, $\pi$, will not affect the diagnostic parameter, $\lambda$, on the same link, and vice versa. The two are orthogonal to each other since they depend on two disjoint sets of data. Therefore, no feedback or "circular reasoning" can take place -- any perturbation of beliefs due to new evidence propagates through the network and is absorbed at the boundary without reflection, resulting in a new equilibrium state compatible with the newly observed evidence.

In summary, we see that the architectural objectives of propagating beliefs coherently, through an active network of primitive, identical, and autonomous processors can be fully realized in singly-connected graphs. Instabilities due to cyclic inferences are avoided by using multiple, source-identified belief parameters, and equilibrium is guaranteed to be reached in time proportional to the network diameter.

The primitive processors are simple, repetitive, and require no working memory except that used in matrix multiplications. Thus, this architecture lends itself naturally to hardware implementation, capable of real-time interpretation of rapidly changing data. It also provides a reasonable model of neural nets involved in cognitive tasks such as visual recognition, reading comprehension [Rumelhart, 1976], and associative retrieval [Anderson, 1983], where unsupervised concurrent processing is an uncontested mechanism.

## 3. PROPAGATION IN MULTIPLY-CONNECTED NETWORKS

The efficacy of singly-connected networks in supporting autonomous propagation raises the question of whether similar propagation mechanisms exist in less restrictive networks (like the one in Figure 1), where multiple parents of common children also possess common ancestors, thus forming loops in the underlying network. If we ignore the existence of loops and permit the nodes to continue communicating with each other as if the network was singly-connected, it will set up messages circulating indefinitely around the loops and the process most probably will not converge to a coherent equilibrium.

A straightforward way of handling the network of Figure 1 would be to appoint a local interpreter for the loop $x_1, x_2, x_3, x_5$ that will pass messages directly between $x_1$ and $x_5$, accounting for the interactions between $x_2$ and $x_3$. This amounts basically to collapsing nodes $x_2$ and $x_3$ into a single node, representing the compound variable $(x_1, x_2)$. This method works well on small loops, but as soon as the number of variables exceeds 3 or 4, collapsing requires handling huge matrices and washes away the natural conceptual structure embedded in the original network.

A second method of propagation is based on "stochastic relaxation" [Geman and Geman, 1984]. Each processor interrogates the states of the variables within its influencing neighborhood, computes a belief distribution for the values of its host variable, then randomly selects one of these values with probability given by the computed distribution. The value chosen will subsequently be interrogated by the neighbors upon computing their beliefs, and so on. This scheme is guaranteed convergence, but usually requires very long relaxation times to reach a steady state.

## Propagation by Conditioning

The method that we found most promising is based on the ability to change the connectivity of a network, and render it singly connected by instantiating a selected group of variables. In Figure 1, for example, instantiating $x_1$ to some value would block the pathway $x_2, x_1, x_3$ and would render the rest of the network singly connected, where the propagation techniques of the preceding section are applicable. Thus, if we wish to propagate the impact of an observed data, say at $x_6$, to the entire network, we first assume $x_1 = 0$, propagate the impact of $x_6$ to the variables $x_2, \ldots, x_5$, repeat the propagation under the assumption $x_1 = 1$ and, finally, linearly combine the two results weighed by the prior probability $P(x_1)$.

The legitimacy of this method is clearly seen from the ever-so-faithful conditioning rule of probability:

$$P(x_2, \cdots x_6) = \sum_{x_1} P(x_2, \ldots, x_6 \mid x_1) \, P(x_1)$$

The novelty here is that while in the ordinary use of the rule we seek a conditioning variable that renders some other variables independent (separating the network into unconnected fragments), we now settle for more modest goals, requiring only that the resulting conditional probability $P(x_2, \ldots, x_6 | x_1)$ have a singly-connected network representation. Note that the choice of $x_2$ as a conditioning variable would be equally adequate, but $x_5$ is a bad choice, since instantiating this variable would not disconnect the pathway $x_2, x_5, x_3$.

The tool of conditioning is not foreign to human reasoning. The terms "hypothetical" or "assumption-based" reasoning, "reasoning by cases," and "envisioning" all refer to the same basic mechanism of selecting a key variable, binding it to some of its values, deriving the consequences of each binding separately, and integrating those consequences together. Reasoning by cases is very frequently used in explanation and argumentation -- showing that diametrically opposed assumptions impart equal credence to a given proposition constitutes a convincing argument for assigning that credence to the proposition. Likewise, showing that different sets of circumstances would require the same type of action constitutes a strong argument for recommending that action.

Although conditioning was introduced here as a sequential process, it can easily be implemented in parallel, to comply with our propagation paradigm. Instead of a single set of $\pi$ - $\lambda$ parameters, each node should maintain several such sets, one for each value of the conditioning variable. The constraint equations (Eqs. (10) and (11)) are checked for each of these sets individually, and the appropriate parameters updated. Additionally, the prior probability of the conditioning variable can also pass along from node to node so that when the overall belief in a given proposition is required, the proper weights will be available to perform the averaging.

## An Illustration

As an example, consider the network in Figure 1 and assume that all variables are binary. Under ordinary updating conditions, with the loops ignored, nodes $x_2, x_3$ and $x_4$ would receive from node $x_1$ the parameters $\pi_{x_2}(x_1) = \pi_{x_4}(x_1) = \pi_{x_3}(x_1) = P(x_1)$, $x_1 = 0,1$, since initially, all $\lambda$'s are set to $(1,1)$. Subsequently, $x_2$ and $x_3$ will compute for their children the parameters $\pi_{x_4}(x_2), \pi_{x_5}(x_2)$ and $\pi_{x_5}(x_3)$ where, using Eq. (11),

$$\pi_{x_5}(x_3) = \sum_{i=0,1} P(x_3 \mid x_1 = i)\, P(x_1 = i)$$

$$\pi_{x_4}(x_2) = \pi_{x_5}(x_2) = \sum_{i=0,1} P(x_2 \mid x_1 = i)\, P(x_1 = i)$$

If these parameters would later be used by $x_5$ and $x_4$ for computing their belief distributions, erroneous quantities would result because the parents are not mutually independent.

By contrast, under conditioning routines, node $x_2$ (as well as $x_3$) will prepare for $x_5$, not a single parameter $\pi_{x_5}(x_2)$, but two:

$$\pi_{x_5}^0(x_2) = P(x_2 \mid x_1 = 0) \qquad x_2 = 0,1$$

$$\pi_{x_5}^1(x_2) = P(x_2 \mid x_1 = 1) \qquad x_2 = 0,1$$

together with the prior probability $P(x_1)$ (see Figure 3). Receiving these two forces $x_5$ to follow suit and compute two sets of parameters as well:

$$\pi_{x_6}^0(x_5) = \sum_{x_2 x_3 = 0,1} P(x_5 \mid x_2 x_3)\pi_{x_5}^0(x_2)\pi_{x_5}^0(x_3)$$

$$\pi_{x_6}^1(x_5) = \sum_{x_2 x_3 = 0,1} P(x_5 \mid x_2 x_3)\pi_{x_5}^1(x_2)\pi_{x_5}^1(x_3)$$

Now imagine that some evidence is obtained, say $x_6 = 1$. Node $x_6$ will provide $x_5$ with the diagnostic parameters:

$$\lambda_{x_6}(x_5) = P(x_6 = 1 \mid x_5) \quad x_5 = 0,1$$

and subsequently, $x_5$ will deliver to $x_2$ two sets of $\lambda_{x_5}(x_2)$ parameters:

$$\lambda_{x_5}^0(x_2) = \alpha^0 \sum_{x_3} \pi_{x_5}^0(x_3) \sum_{x_5} P(x_5 \mid x_2 x_3)\lambda_{x_6}(x_5)$$

$$\lambda_{x_5}^1(x_2) = \alpha^1 \sum_{x_3} \pi_{x_5}^1(x_3) \sum_{x_5} P(x_5 \mid x_2 x_3)\lambda_{x_6}(x_5)$$

A similar set will be computed for $x_3$. To calculate the overall belief in a proposition, say $x_2$, we make use of the prior probability $P(x_1)$ and compute the average

$$BEL\,(x_2) = \beta^0\, \pi_{x_5}^0(x_2)\, \lambda_{x_5}^0(x_2)\, P(x_1 = 0) + \beta^1\, \pi_{x_5}^1(x_2)\, \lambda_{x_5}^1(x_2)\, P(x_1 = 1)$$

It is essential to note that the conditioned parameters must propagate as separate quantities and only be averaged when the final belief measures are to be calculated. The reason is that the conditioning variable influences the other variables in the loop along two separate paths, clockwise and counterclockwise. If we were to pass along the averaged quantities, instead of the individual constituents, it would amount to counting the prior information twice, instead of once.

## CONCLUSIONS

The architectural objectives of propagating beliefs coherently by self-activated and concurrent mechanisms are fully realizable in singly-connected graphs. In multiply-connected graphs the mechanism can still be applicable but requires duplicating the number of messages for each conditioning variable.

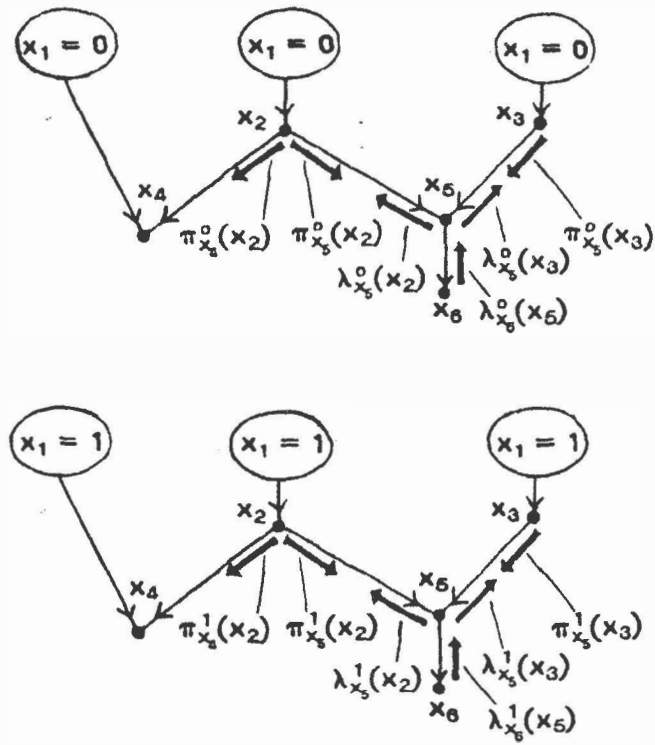

*Figure 3*

While the removal of a single node is sufficient to break up all loops in the network of Figure 1, more complex networks may require several such nodes. When this happens, the propagation must be conditioned on all value combinations of the variables in this cutset, and their number might be substantial. It is important, therefore, to find as small a cutset as possible. Although the problem of finding a minimal cutset is probably NP hard, simple heuristics exist for finding close-to-optimal sets [Levy and Low, 1983]. Moreover, the effort invested in searching for a small cutset will be amortized over many propagation instances, as long as the network topology remains the same.

## REFERENCES

Anderson, John R., (1983), "The Architecture of Cognition", Harvard University Press, Cambridge, MA.

Duda, R. O., Hart, P. E., and Nilsson, N. J., (1976), "Subjective Bayesian Methods for Rule-

Based Inference Systems", in *Proc. 1976 National Computer Conference (AFIPS Conference Proceedings)*, 45, 1075-1082.

Geman, S. and Geman, D., (1984), Stochastic Relaxations, Gibbs Distributions, and the Bayesian Restoration of Images, *IEEE Trans. on Pattern Analysis and Machine Intelligence*, PAMI-6, No. 6, 721-742, November.

Howard, R.A., and Matheson, J.E, (1984), "Reading on the Principles and Applications of Decision Analysis", Strategic Decisions Group, Menlo Park, CA.

Kim, J. and Pearl, J., (1983), "A Computational Model for Combined Causal and Diagnostic Reasoning in Inference Systems", *Proceedings of IJCAI-83*, 190-193.

Levy, H. and Low, D. W. (1983), "A New Algorithm for Finding Small Cycle Cutsets," Report G 320-2721, IBM Los Angeles Scientific Center.

Lowrance, J. D., (1982), "Dependency-Graph Models of Evidential Support", COINS Technical Report 82-26, University of Massachusetts at Amherst.

McAllester, D., (1980), "An Outlook on Truth Maintenance", Artificial Intelligence Laboratory, AIM-551, Cambridge: MIT.

Pearl, J., (1982), "Reverend Bayes on Inference Engines: A Distributed Hierarchical Approach", in *Proc. of AAAI Conference on Artificial Intelligence*, Pittsburgh, PA, 133-136.

Pearl, J., (1985), "Bayes and Markov Networks, A Comparison of Two Graphical Representations of Probabilistic Knowledge", Cognitive Systems Laboratory, Technical Report R-46, in preparation.

Rosenfeld, A., Hummel, A., and Zucker, S., (1976), "Scene Labelling by Relaxation Operations", *IEEE Trans. on Computers*, pp. 562-569.

Rumelhart, D. E., (1976), toward an Interactive Model of Reading, *Center for Human Info. Proc. CHIP-56*, UC San Diego, La Jolla, CA.

Waltz, D. G., (1972), "Generating Semantic Descriptions from Drawings of Scenes with Shadows", AI TR-271, Artificial Intelligence Laboratory, Massachusetts Institute of Technology, Cambridge, MA.